\documentclass[10pt,twocolumn,letterpaper]{article}

\usepackage{iccv}
\usepackage{times}
\usepackage{epsfig}
\usepackage{graphicx}
\usepackage{amsmath}
\usepackage{amssymb}
\usepackage{multirow}
\usepackage{wrapfig}

\usepackage{subfigure}
\usepackage{color}

\usepackage[pagebackref=true,breaklinks=true,letterpaper=true,colorlinks,bookmarks=false]{hyperref}

\iccvfinalcopy 

\def\iccvPaperID{****} 

\ificcvfinal\fi

\begin{document}

\title{TruNet: Short Videos Generation from Long Videos via Story-Preserving Truncation}

\author{Fan Yang$^3$~\thanks{This work was done when Fan Yang was a FTE at Baidu.}\quad
Xiao Liu~$^{1}$ \quad
Dongliang He~$^{1}$\quad
Chuang Gan~$^{2}$\quad\\
Jian Wang~$^{1}$\quad
Chao Li~$^{1}$\quad
Fu Li~$^{1}$\quad
Shilei Wen~$^{1}$ \\
$^1$ Department of Computer Vision Technology (VIS), Baidu Inc., China\\
$^2$ MIT-Watson AI Lab \\
$^3$ Donald Bren School of Information and Computer Sciences, University of California, Irvine\\
{\tt\small fyang7@uci.edu~~~~ganchuang1990@gmail.com}\\
{\tt\small \{liuxiao12,hedongliang01,wangjian33,lifu,wenshilei\}@baidu.com}
}

\maketitle
\ificcvfinal\thispagestyle{empty}\fi

\begin{abstract}
In this work, we introduce a new problem, named as {\em story-preserving long video truncation}, that requires an algorithm to automatically truncate a long-duration video into multiple short and attractive sub-videos with each one containing an unbroken story.
This differs from traditional video highlight detection or video summarization problems in that each sub-video is required to maintain a coherent and integral story,
which is becoming particularly important for resource-production video sharing platforms such as Youtube, Facebook, TikTok, Kwai, etc.
To address the problem, we collect and annotate a new large video truncation dataset, named as {\em TruNet}, which contains 1470 videos with on average 11 short stories per video.
With the new dataset, we further develop and train a neural architecture for video truncation that consists of two components: a Boundary Aware Network (BAN) and a Fast-Forward Long Short-Term Memory (FF-LSTM).
We first use the BAN to generate high quality temporal proposals by jointly considering frame-level attractiveness and boundaryness.
We then apply the FF-LSTM, which tends to capture high-order dependencies among a sequence of frames, to decide whether a temporal proposal is a coherent and integral story.
We show that our proposed framework outperforms existing approaches for the story-preserving long video truncation problem in both quantitative measures and user-study.
The dataset is available for public academic research usage at \href{https://ai.baidu.com/broad/download}{https://ai.baidu.com/broad/download}.
\end{abstract}

\begin{figure}[t]
	\begin{center}
		\subfigure[An variety show with 9 song and dance performances. ]{
			\includegraphics[scale = 0.26]{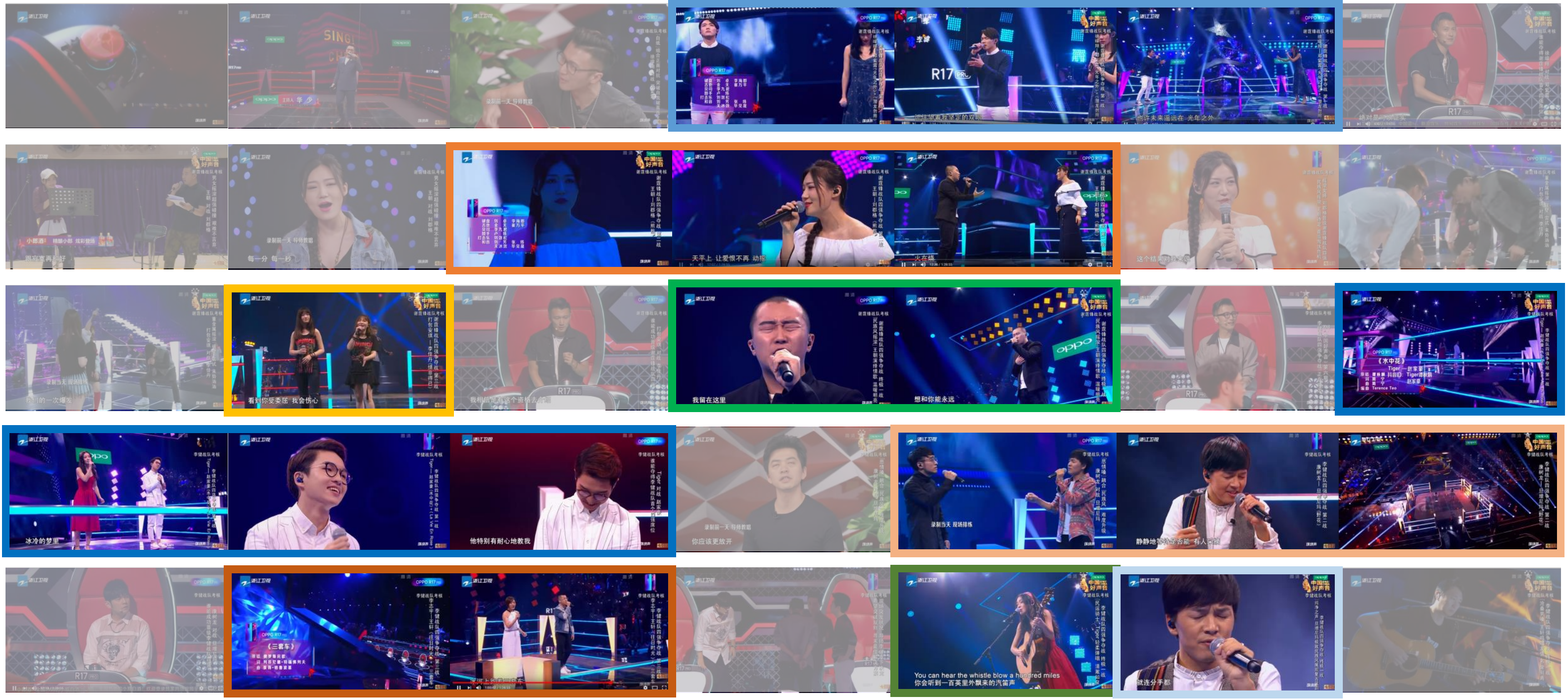}
		}
		\subfigure[The third performance with a higher temporal resolution. ]{
			\includegraphics[scale = 0.3]{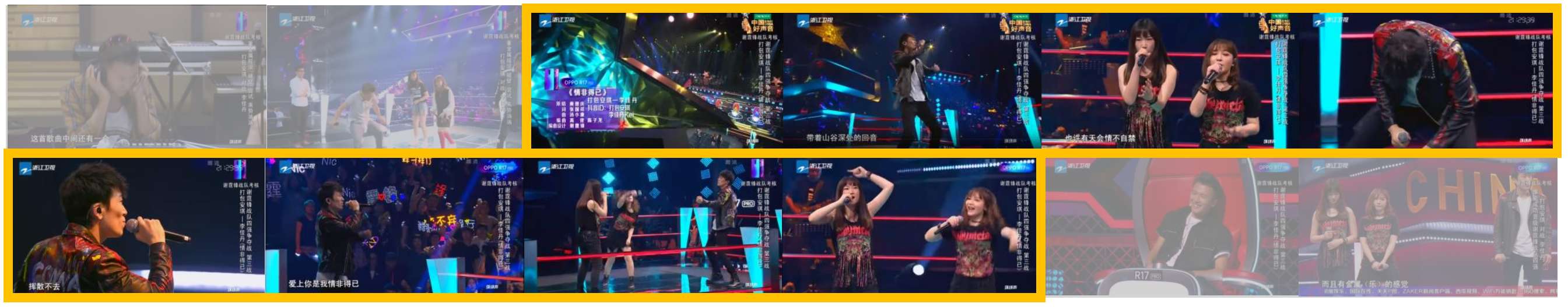}
		}
	\end{center}
	\vspace{-10pt}
	\caption{TruNet provides on average 11 short stories per long video.
A sample video of the TruNet is shown in (a). The video is a variety show that contains 9 song and dance performances.
The 9 short stories are indicated in different colors, and the third one is shown in (b) with a higher temporal resolution.
    }
	\label{fig:title}
\end{figure}

\section{Introduction}
Short-form video sharing platforms such as TikTok and Kwai are becoming increasingly popular and lead to the requirement of generating short-form videos.
Users prefer to consume their time in more compact short-form videos, while high quality videos such as reality show and TV series are usually long (e.g. $>$ 1 hour).
In such context, developing new algorithms that can truncate long videos into short, attractive and unbroken stories is of special interest.

\begin{table*}[htb]
	\begin{center}
		\begin{tabular}
			{c||c||c}\hline
			& ActivityNet1.3 & TruNet \\\hline\hline
			Total Video Number & 19994 & 1470 \\\hline
			Total Video Length & 648.2h & 2101.0h \\\hline
			Average Video Length & 2.0min & 80.0min \\\hline
			Total Story(Action) Length & 315.3h & 845h \\\hline
			Total Story(Action) Number & 23064 & 16891 \\\hline
			Average Story(Action) Length & 0.8min & 3min \\\hline
			Class Number & 200 & 2 \\\hline
			Average Story(Action) Number per Video & 1.2 & 11.0 \\\hline
		\end{tabular}		
	\end{center}
	\vspace{-10pt}
	\caption{Comparison between ActivityNet1.3~\cite{Heilbron2015activitynet} and TruNet. Both provide large scale data annotation. Although ActivityNet contains more videos, the average video length and total story length in TruNet are significantly higher.}
	\label{tab:data_summary}
\end{table*}

Although much progress has been made in video highlight detection and video summarization \cite{yang2015highlight,yao2016highlight,gygli2014highlight,gygli2015video,goldman2014storyboard}, many of the them focus on producing a coherent whole story in the final combined highlight and there is no requirement of an unbroken story to each sub-video.
And the problem of story-preserving long video truncation is still not well studied due to the limitation of existing datasets.
Obviously, story integrity or completeness is a crucial measurement for short-form videos under this scenario, but is not yet considered in existing video highlight datasets.
For example, a climactic fight fragment is interesting and important enough to be a ground-truth keyshot in SumMe \cite{gygli2014highlight} and TVSum \cite{song2015dataset},
but as an individual short-form video, it has to involve the beginning and the ending parts to clarify the cause and effect of the story.
On the other hand, ActivityNet \cite{Heilbron2015activitynet} provides action intervals with accurate temporal boundaries, but its data distribution is different from the requirement of short-form video production from the massive long video database.
As Table 1 shows, the average video length of ActivityNet is too short ($<$ 2 minutes) such that the average story number per video is only 1.2 and the average story length is only 0.8 minute.

In this paper, we collect a new large dataset, named as TruNet, which contains 1470 videos with a total of 2101 hours and an average video length longer than 1 hour.
It covers a wide range of popular topics including variety show, reality show, talk show, and TV series.
TruNet provides on average 11 short stories per long video and each story is annotated with accurate temporal boundaries.
Figure 1(a) shows A sample video of the TruNet. The video is a variety show that contains 9 song and dance performances.
The 9 short stories are indicated in different colors, and the third one is shown in Figure 1(b) with a higher temporal resolution.

With the new dataset, we further develop a baseline neural architecture for story-preserving long video truncation that consists of two components: a Boundary Aware Network (BAN) for proposal generation and a Fast-Forward Long Short-Term Memory (FF-LSTM) for story integrity classification.
Different from previous state-of-the-art methods \cite{xiong2017tag} whose proposal generation only depends on actionness, BAN utilizes additional frame-level boundaryness to generate proposals and achieves higher precision when the number of proposals is small.
And different from traditional LSTM on sequence modeling, FF-LSTM \cite{zhou2016ff} introduces fast-forward connections to a stack of LSTM layers to encourage stable and effective backpropagation in the deep recurrent topology,
which leads to an obvious performance improvement in our video story classification.
To the best of our knowledge, this is the first time that FF-LSTM has been used for modeling sequences in the video domain.



In summary, our contributions are threefold:
(1) We introduce a new practical problem in video truncation, {\em story-preserving long video truncation}, which requires to truncate a long-time video into multiple short-form videos with each one preserving a story.
(2) We collect and annotate a new large dataset for studying this problem, which can become a complementary source to existing video datasets.
(3) We propose a baseline framework that involves a new temporal proposal generation module and a new sequence modeling module, with better performance compared to traditional methods.
We will also release the dataset for public academic research usage.

\section{Related Work}
\subsection{Video Dataset}
Here we briefly review typical video datasets that are related to our work.
The SumMe dataset \cite{gygli2014highlight} consists of 25 videos covering 3 categories.
The length of the videos ranges from about 1 to 6 minutes and $5\%$ to $15\%$ frames are extracted to be the summary of a video.
Similarly, the TVSum dataset \cite{song2015dataset} contains 50 videos from 10 categories.
The video duration is between 2 to 10 minutes and at most $5\%$ frames are selected to be the keyshots.
Compared with SueMe and TVSum, our proposed TruNet dataset focuses on the story-preserving long video truncation problem such that each summary is an integral short-form video with accurate temporal boundaries.

On the other hand, although THUMOS14 \cite{jiang204thumos} and ActivityNet \cite{Heilbron2015activitynet} also provide temporal boundaries,
their data distributions are not suitable for our problem.
Take ActivityNet as an example, the average video length of ActivityNet is too short and the average action number per video is only 1.2.
Whether the daily activities collected by ActivityNet are suitable for video sharing is another consideration.
In contrast, our proposed TruNet focuses on collecting high-quality long videos,
such that the average short story number per video can be up to 11.
The video topics in TruNet are also chosen to be suitable to share in video sharing platforms.

\subsection{Video Summarization}
Video summarization has been a long-standing problem in computer vision and multimedia.
Previous studies typically treat it as extracting keyshots \cite{yang2015highlight,Cernekov2006keyframe,lee2012highlight,wolf1996keyframe,khosla2013keyframe,kim2013keyframe}, video skims \cite{yao2016highlight,xu2015gaze,zhao2014skim,gygli2014highlight,gygli2015video,lu2013story,potapov2014skim}, storyboards \cite{goldman2014storyboard}, time-lapses \cite{kopf2014summary}, montages \cite{sun2014montage}, and video synopses \cite{pritch2008summary}.
An exhaustive review is beyond the scope of this paper. We refer \cite{truong2007survey} for a survey of early works on video summarization.
Compared with previous approaches, this paper proposes story-preserving long video truncation that formulates the video summarization problem in a different way.
Each truncated short-form video should be an attractive and unbroken short story such that it can be used in the video sharing platforms.

\subsection{Temporal Action Localization}
Our work is also related to the work on temporal action localization.
Early methods \cite{tang2013localization,jain2014localization,yuan2016localization} of temporal action location relied on hand-crafted features and sliding window search.
Using ConvNet-based features such as C3D \cite{shou2016temporal} and two-stream CNN \cite{dai2017temporal,yuan2017temporal} achieves both higher efficiency and better performance than hand-crafted features.
\cite{escorcia2016temporal,gao2017turn,xiong2017tag} focused on utilizing stronger network structure to generate high-quality temporal proposals.
Structured temporal modeling \cite{zhao2017temporal} and 1d temporal convolution \cite{lin2017single} are also proved important to boost the performance of temporal action localization.
While conceptually similar, our proposed model is novel in that it generates high quality temporal proposals by jointly considering frame-level attractiveness and boundaryness,
which facilitates the truncated short-form videos to be unbroken. BSN \cite{lin2018bsn} is also boundary sentitive framework, but it is designed for temporal action proposal and its proposals are generated from frame-level action-start and action-end confidence score, our BAN generate proposals according to story-start, story-end as well as the storyness score at group-of-frames level to suite the extremely long input.

\section{The TruNet Dataset}

\begin{figure}[!t]
 	\begin{center}
		\includegraphics[scale = 0.4]{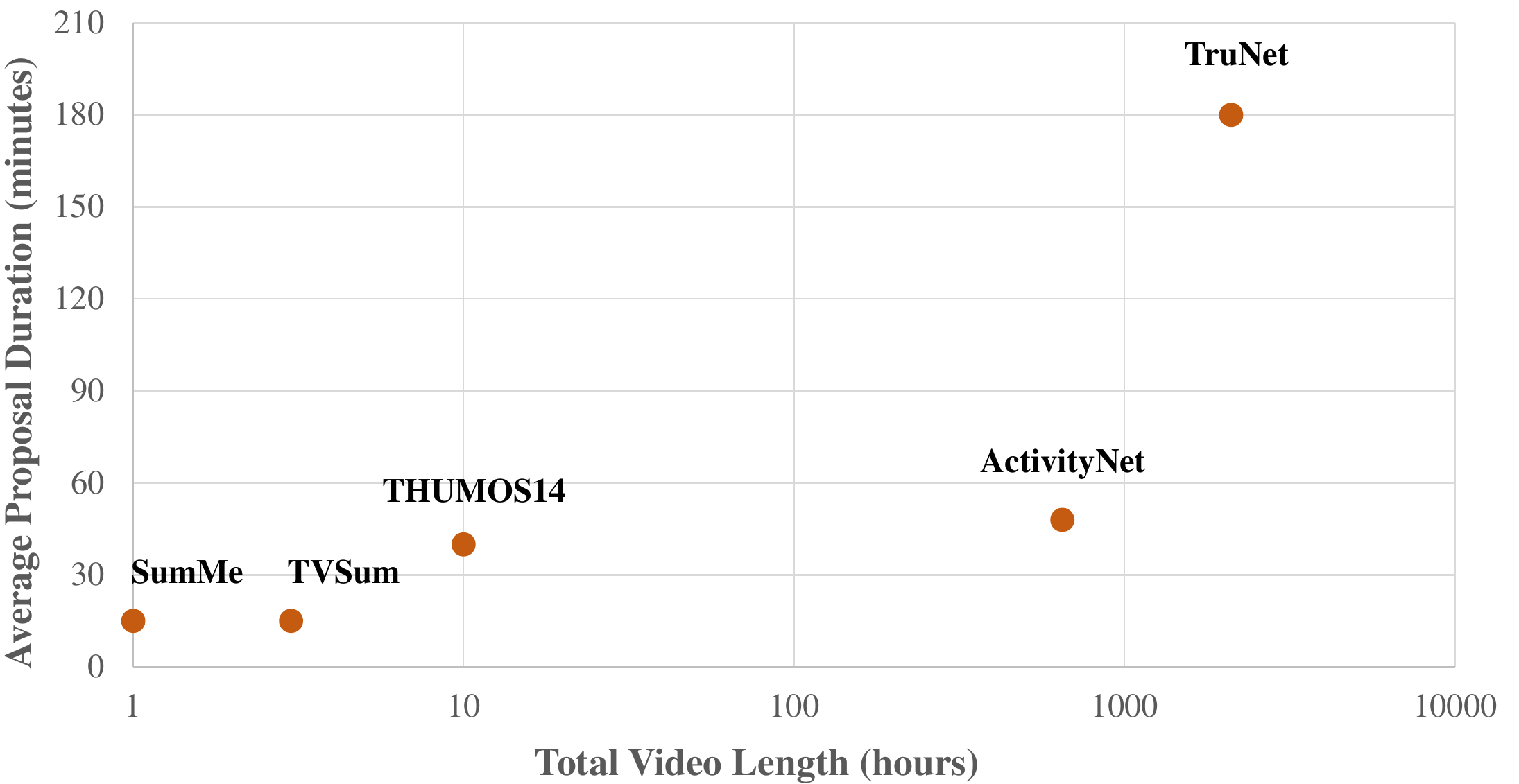}
	\end{center}
	\vspace{-10pt}
	\caption{Compared to existing video summarization and temporal action localization datasets, our TruNet dataset has the longest total video length and average proposal duration.}
	\label{fig:fig4}
\end{figure}

\begin{figure}[t]
	\begin{center}
		\subfigure[Video length distribution. ]{
			\includegraphics[scale = 0.25]{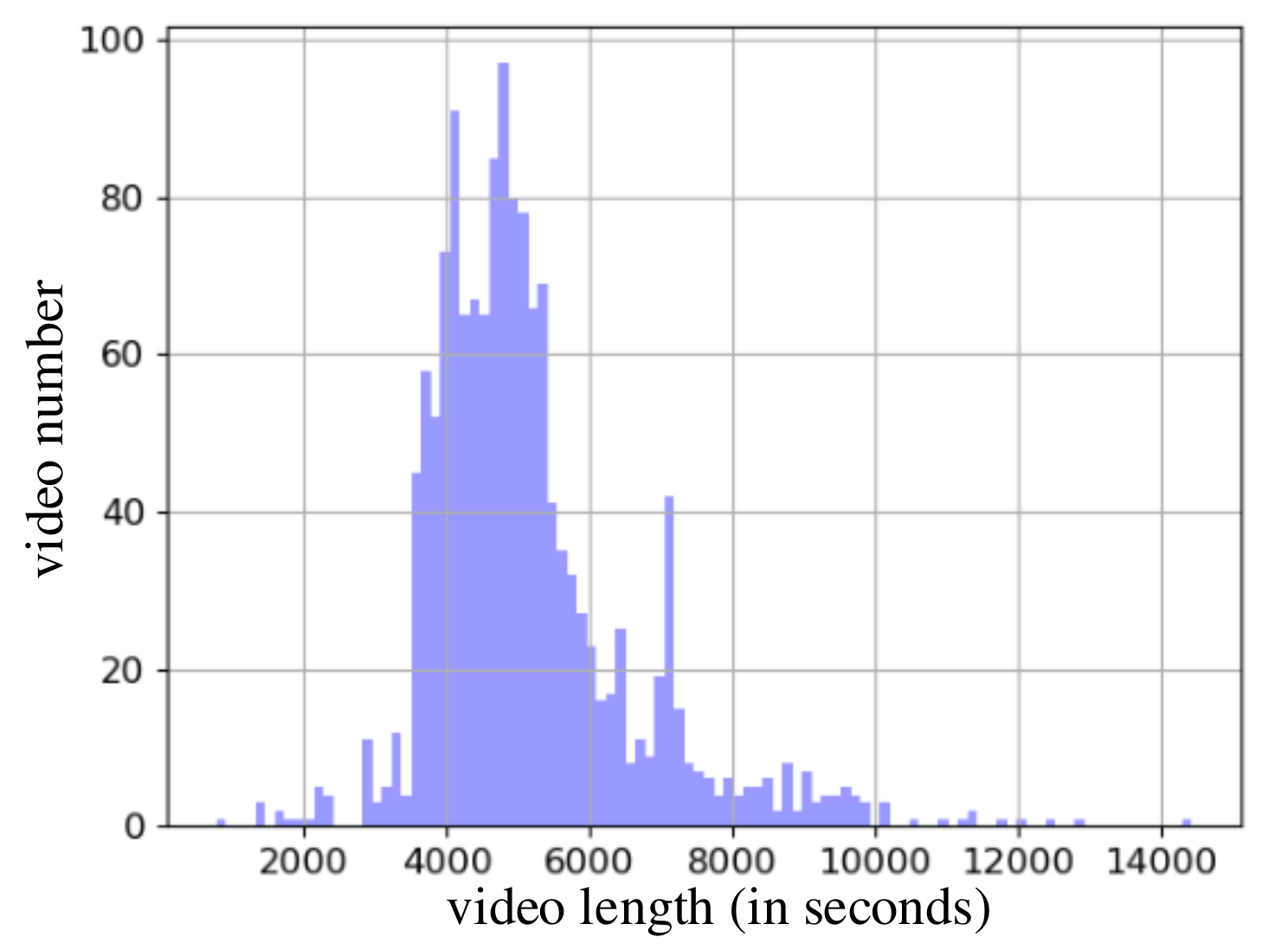}
		}
		\subfigure[Story number distribution. ]{
			\includegraphics[scale = 0.25]{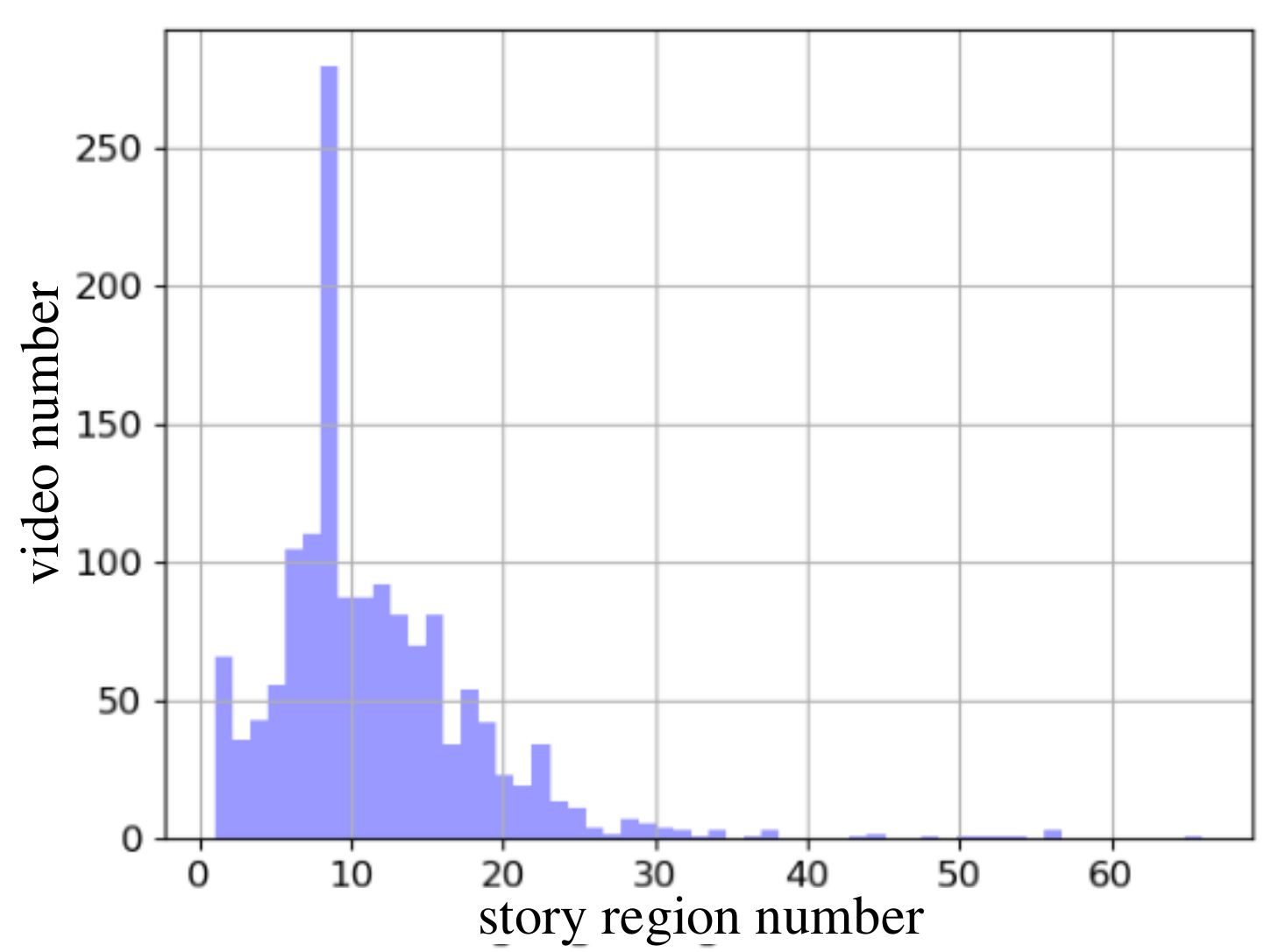}
		}
		\subfigure[Story length distribution. ]{
			\includegraphics[scale = 0.25]{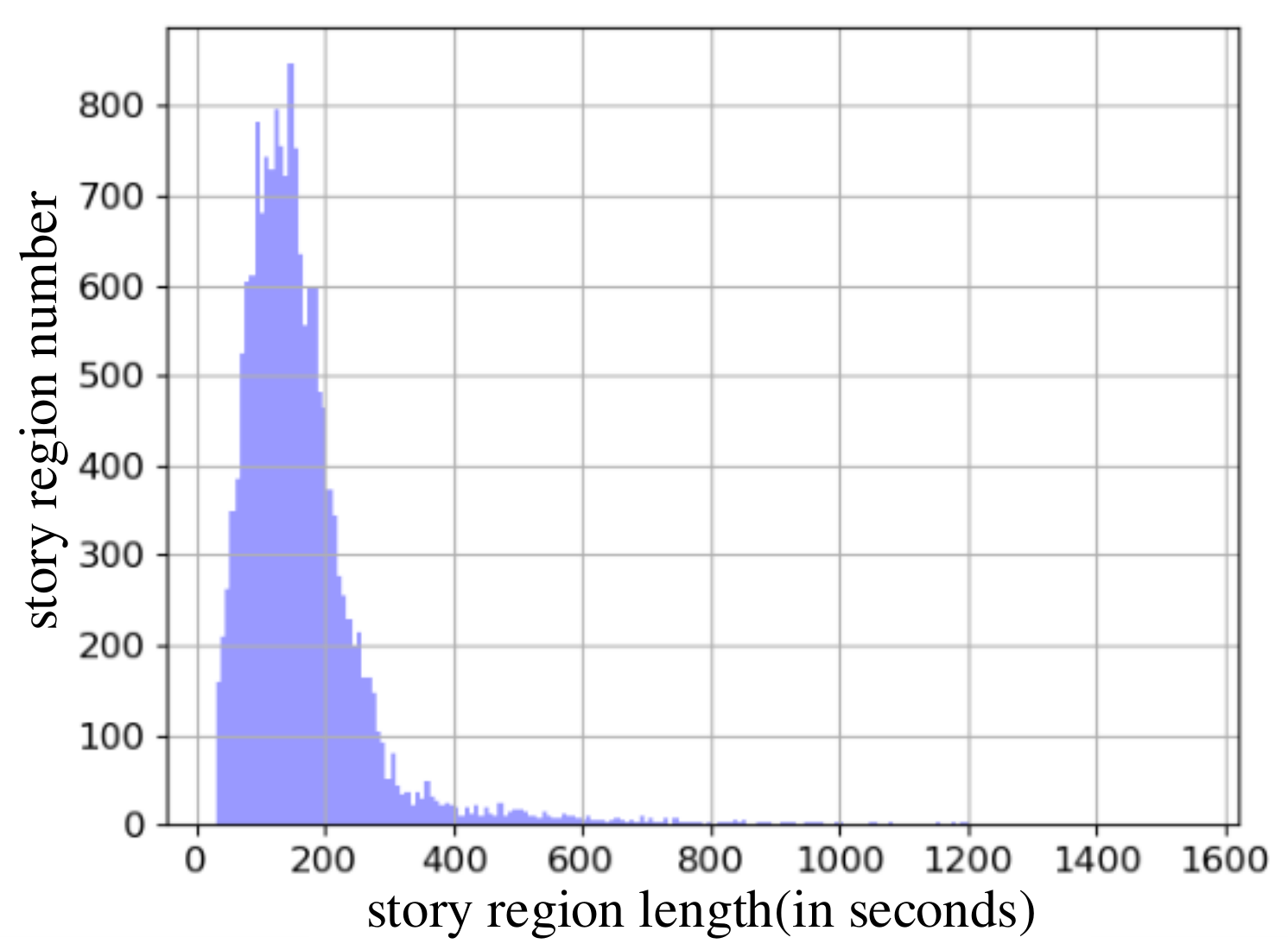}
		}
		\subfigure[Story ratio distribution. ]{
			\includegraphics[scale = 0.25]{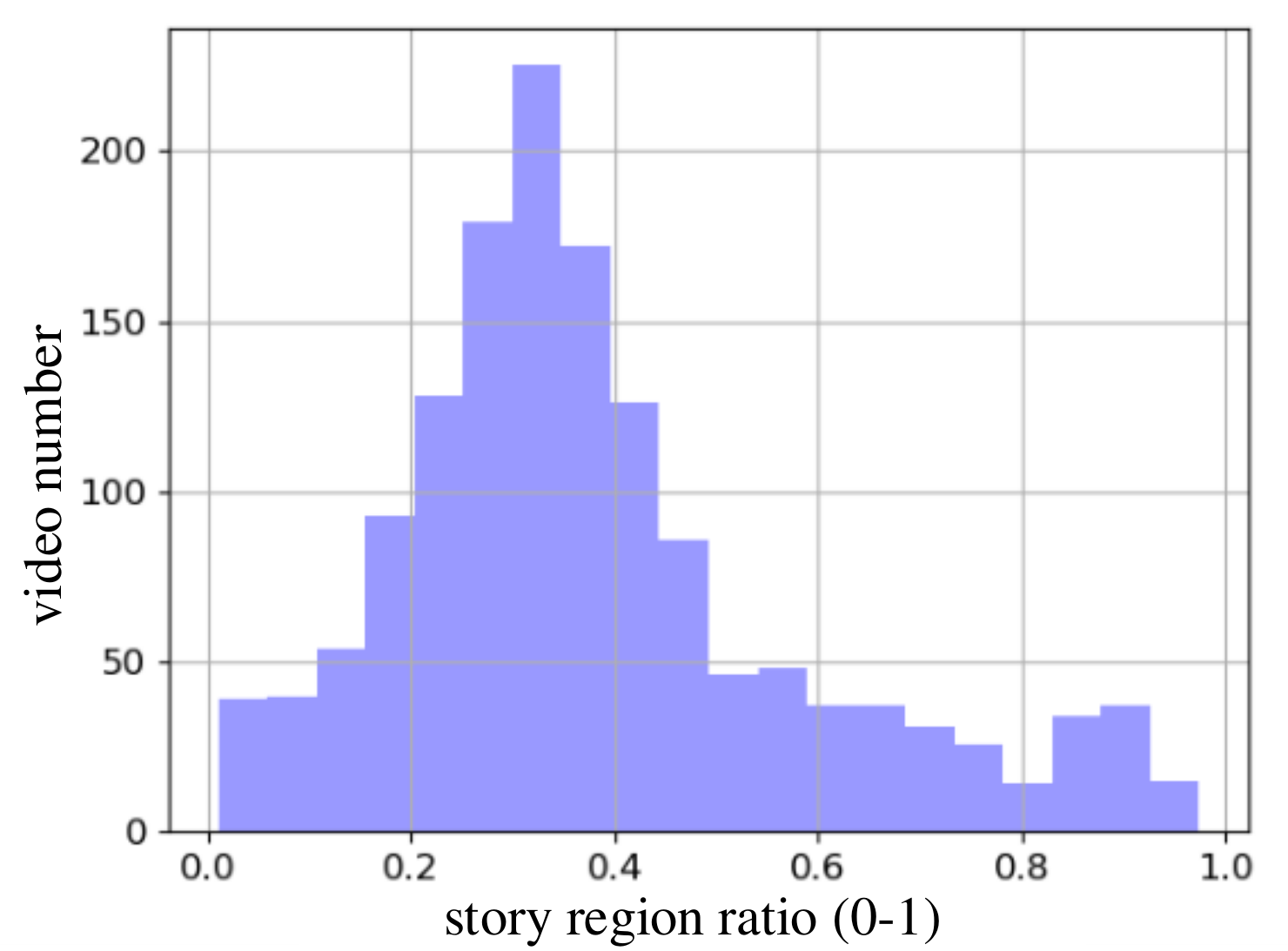}
		}
	\end{center}
	\vspace{-10pt}
	\caption{Statistics of the TruNet Dataset. }
	\label{fig:fig5}
\end{figure}

The story-preserving long video truncation problem is not well studied due to the lack of publicly available dataset.
Therefore, we construct the TruNet dataset to quantitatively  evaluate the proposed framework.

\subsection{Dataset Setup}
Considering the truncated short-form stories from long videos should be suitable to share in video sharing platforms,
we choose four types of long videos, \emph{i.e.} variety show, reality show, talk show and TV series.
Most videos of the TruNet are downloaded from the video website: iQIYI.com, which has a large quantity of high quality long videos.
Crowdsourced annotation based on a carefully designed annotation tool is then applied after data collection.
Each annotation worker is trained by annotating a small number of videos, and can participate the formal annotation only when he/she passes the training program.
During the annotation task, the worker is asked to
1) watch the whole video;
2) annotate temporal boundaries of short stories;
3) adjust the boundaries.
Each long videos are annotated by multiple workers for quality assurance, and the annotations are finally reviewed by several experts.
We randomly split the dataset into three parts: training set with 1241 videos, validation set with 115 samples and testing set with 114 samples.
The validation set are not used throughout our experiments.

\subsection{Dataset Statistics}
The TruNet dataset consists of 1470 long videos with the duration of 80 minutes on average and 2101 hours in total.
A long video contains 11 stories on average, and the manually labeled story number is 16891 in total.
The average duration of a short story is 3 minutes.
Figure 2 compares TruNet with existing video summarization and temporal action localization dataset such as SumMe, TVSum, ActivityNet and  THUMOS14.
As can be seen, TruNet has the largest total video length and average proposal duration.

We show a group of statistics of the TruNet dataset in Figure 3.
The video length distribution is shown in Figure 3(a).
The distribution of story number in each long video is shown in Figure 3(b).
The distribution of story length is shown in Figure 3(c).
The distribution of the ratio story region length to long video length is shown in Figure 3(d).

\section{Methodology}

\begin{figure*}[!t]
\begin{center}
\includegraphics[scale = 0.55]{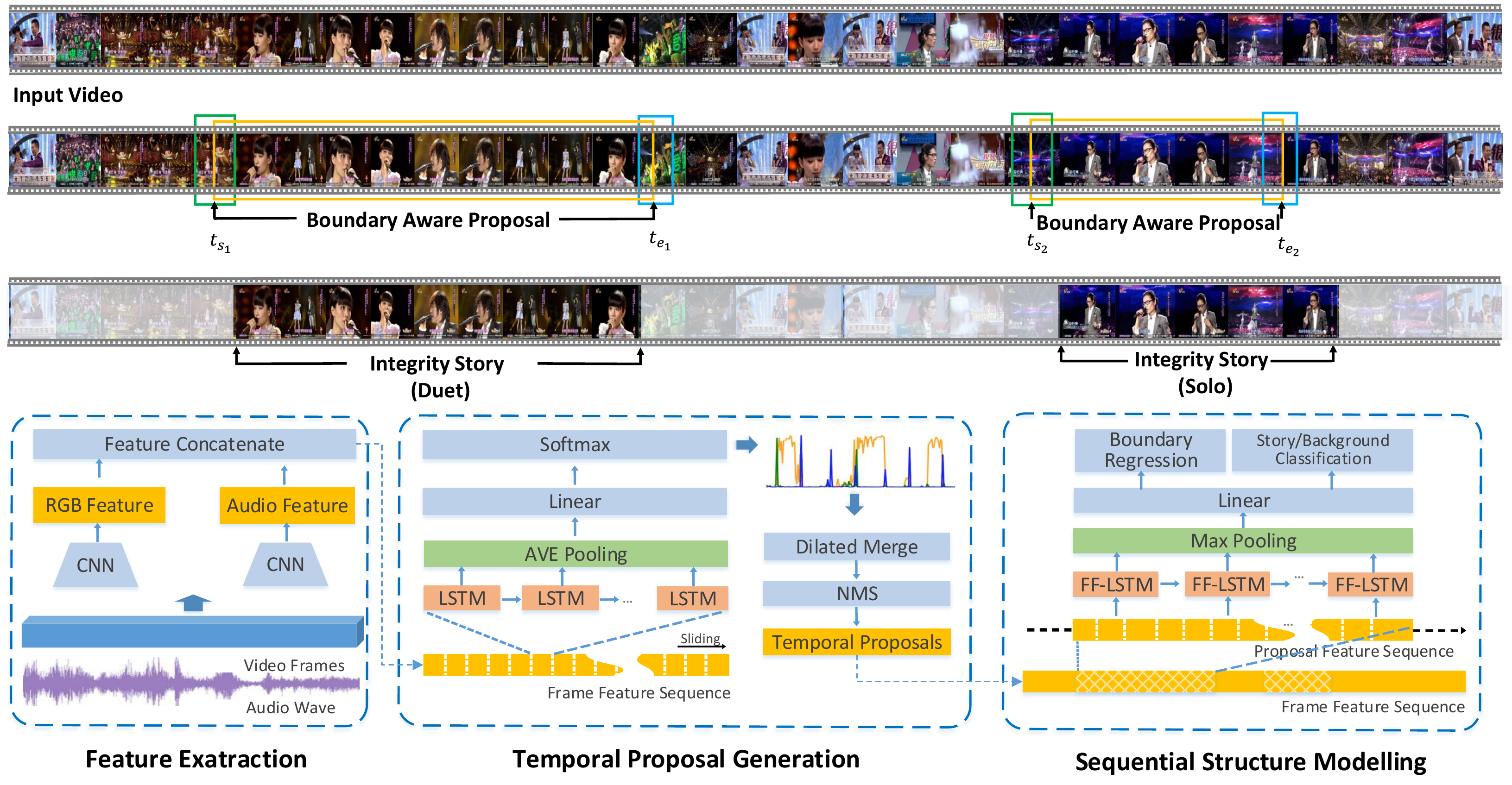}
\end{center}
\vspace{-10pt}
\caption{An architecture overview of the proposed framework which contains three components: feature extraction, temporal proposal generation and sequential structure modeling.
Multi-modal features of frames are extracted and concatenated in the first component.
In the second component, a boundary aware network is used to predict attractiveness and boundaryness for each frame.
A dilated merge algorithm is then carried out to generate proposals.
In the third component, the sequential structure of each proposal is modeled by FF-LSTM, which outputs the classification confidence score together with the refined boundaries.}\label{fig:fig1}
\end{figure*}

The mathematical formulation of the story-preserving long video truncation problem is similar with temporal action localization.
The training dataset can be represented as $\tau=\{V_i = (\{u_t^i\}_{t=1}^{T_i}, \{s_{i,j},e_{i,j}\}_{j=1}^{l_i})\}_{i=1}^N$
where frame-level feature $u_t^i$ comes from long video $V_i$,
which is corresponding with a ground-truth sliced short-form video set.
$\{s_{i,j},e_{i,j}\}$ is the beginning and ending indexes of the $j^{th}$ interval of $V_i$.
$N$ is the number of training videos, $T_i$ is the frame number of $V_i$ and $l_i$ is the interval number of $V_i$.

Figure 4 illustrates the architecture of the proposed framework, which includes three major components: feature extraction, temporal proposal generation and sequential structure modeling.
Given an input long video, RGB-based 2D convolutional feature and audio feature are extracted and concatenated.
In the temporal proposal generation component, the features are fed into the BAN to predict the attractiveness and boundaryness for each frame.
A dilated merge algorithm is then carried out to generate temporal proposals according to the frame-level attractiveness and boundaryness scores.
In the sequential structure modeling component, the sequential structure of each proposal is modeled by FF-LSTM, which outputs the classification confidence score of the proposal together with the refined boundaries.

\subsection{Boundary Aware Network}
A novel boundary aware network (BAN) is proposed in this paper to provide high quality story proposals.
As Figure 4 shows, BAN takes the features of consecutive 7 frames as the input of a LSTM layer.
The output of the LSTM is averaged pooled and a linear layer is utilized to predict a four-categories probability scores, include: within story, background, story beginning boundary and story ending boundary.
The label of the center frame is decided by the category with the largest score.
We regard a frame sequence as a story candidate if every frame in the sequence belongs to the within story category.
A simple dilated merge algorithm is carried out to merge adjacent story candidates with small distance (5 frames).
Non-Maximal Suppression (NMS) is carried out to reduce redundancy and generate the final proposals.

Different from previous popular methods \cite{xiong2017tag,escorcia2016temporal,gao2017turn,zhao2017temporal} that only depend on actionness for proposal generation, BAN utilizes additional frame-level boundaryness to generate proposals.
A comparison between the actionness score of TAG \cite{xiong2017tag} (the upper row) and the proposed BAN (the middle row) is shown in Figure 5.
For TAG, we show actionness scores larger than 0.5.
For BAN, we show the maximum scores among the three foreground categories in different colors.
The ground-truth proposal intervals are shown in the bottom row.
As can be seen, the score curve of BAN is smoother than TAG, and the boundaries better match the ground-truth proposal intervals.

\subsection{FF-LSTM}
We first briefly review LSTM that serves as the basis of sequential structure modeling.
Long-Short Term Memory (LSTM) \cite{hochreiter1997lstm} is an enhanced version of recurrent neural network (RNN) with a set of memory cells $c$. The computation of LSTM can be written as
\begin{eqnarray}
b_i^t &=& \sigma_i (z_i^t)        \nonumber\\
b_g^t &=& \sigma_g (z_g^t + W_gc^{t-1}) \nonumber \\
c_t &=& b_g^tc^{t-1} + b_i^t\sigma_g(z_c) \nonumber \\
b_o^t &=& \sigma_g(z_o + W_oc^{t-1})   \nonumber \\
h^{t} &=& \sigma_o(c^t)b_o^t           \nonumber \\
\left[z_i, z_g, z_c, z_o\right] &=& W_fx^t + W_hh^{t-1}  ,
\end{eqnarray}
where $t$ is the time step, $x$ is the input, $\left[z_i, z_g, z_c, z_o\right]$ is a concatenation of four vectors of equal size, $h$ is the output of $c$,
$b_i$, $b_g$ and $b_o$ are input gate, forget gate and output gate respectively, $\sigma_i$, $\sigma_g$ and $\sigma_o$ are input activation function, forget activation function and output activation function respectively, $W_f$, $W_h$, $W_g$ and $W_o$ are learnable parameters. The computation of (1) can be equivalently split into two consecutive steps: the hidden block  $f^t = W_fx^t$ and the recurrent block $(h^t, c^t) = {\rm LSTM}(f^t, h^{t-1}, c^{t-1})$.

A straightforward deep LSTM can be constructed by directly stacking multiple LSTM layers. Suppose ${\rm LSTM}_k$ is the $k^{th}$ LSTM layer, then:
\begin{eqnarray}
(h^t_k, c^t_k) &=& {\rm LSTM}_k(f^t_k, h^{t-1}_k, c^{t-1}_k)   \nonumber \\
f^t_k &=& W_f^kx^t, \hspace{0.2in}   k = 1   \nonumber \\
f^t_k &=& W_f^kh^t_{k-1}, \hspace{0.1in} k > 1.
\end{eqnarray}

\begin{figure}[!t]
\includegraphics[scale = 0.35]{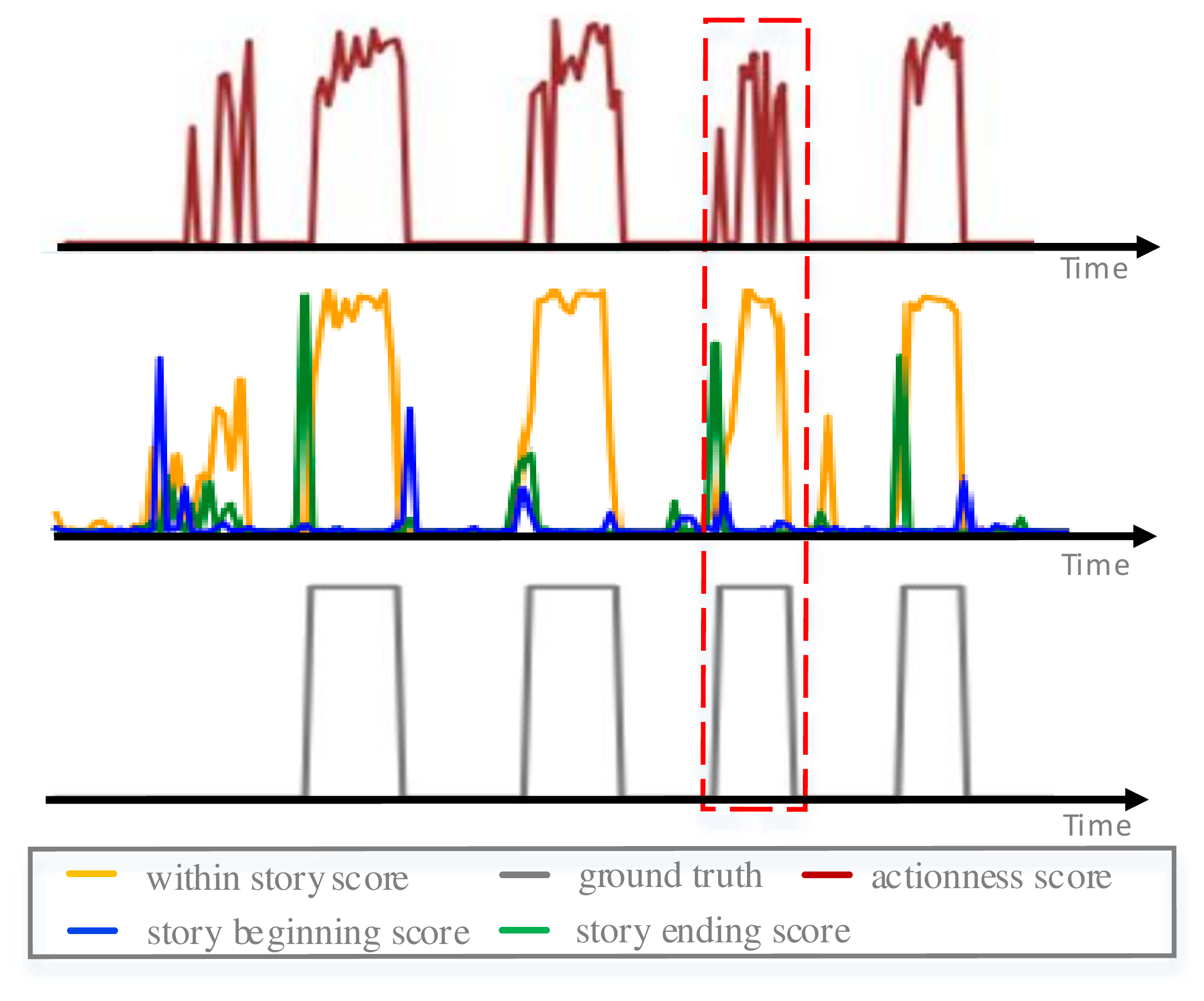}
\vspace{-10pt}
\caption{A comparison between the actionness score of TAG \cite{xiong2017tag} (the upper row) and the proposed BAN (the middle row).
For TAG, we show actionness scores larger than 0.5.
For BAN, we show the maximum scores among the three foreground categories (within story, stroy beginning boundary and story ending boundary) in different colors.
The ground-truth proposal intervals are shown in the bottom row.
}\label{fig:fig3}
\end{figure}

\setlength{\tabcolsep}{1pt}
\begin{figure*}[t]
\begin{center}
\subfigure[FF-LSTM]{\includegraphics[scale=0.6]{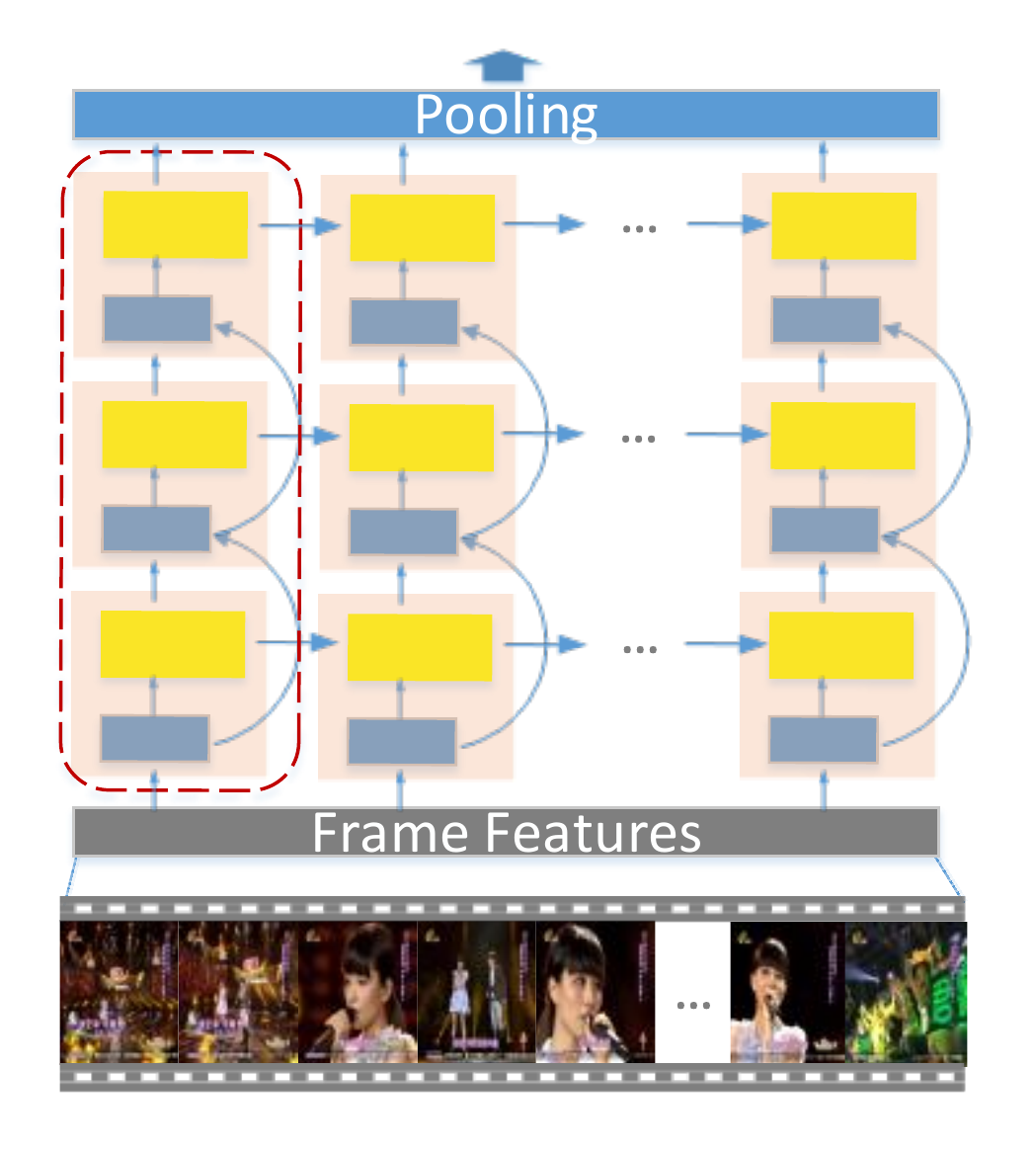}}
\subfigure[LSTM\hspace{0.4in}]{\includegraphics[scale=0.6]{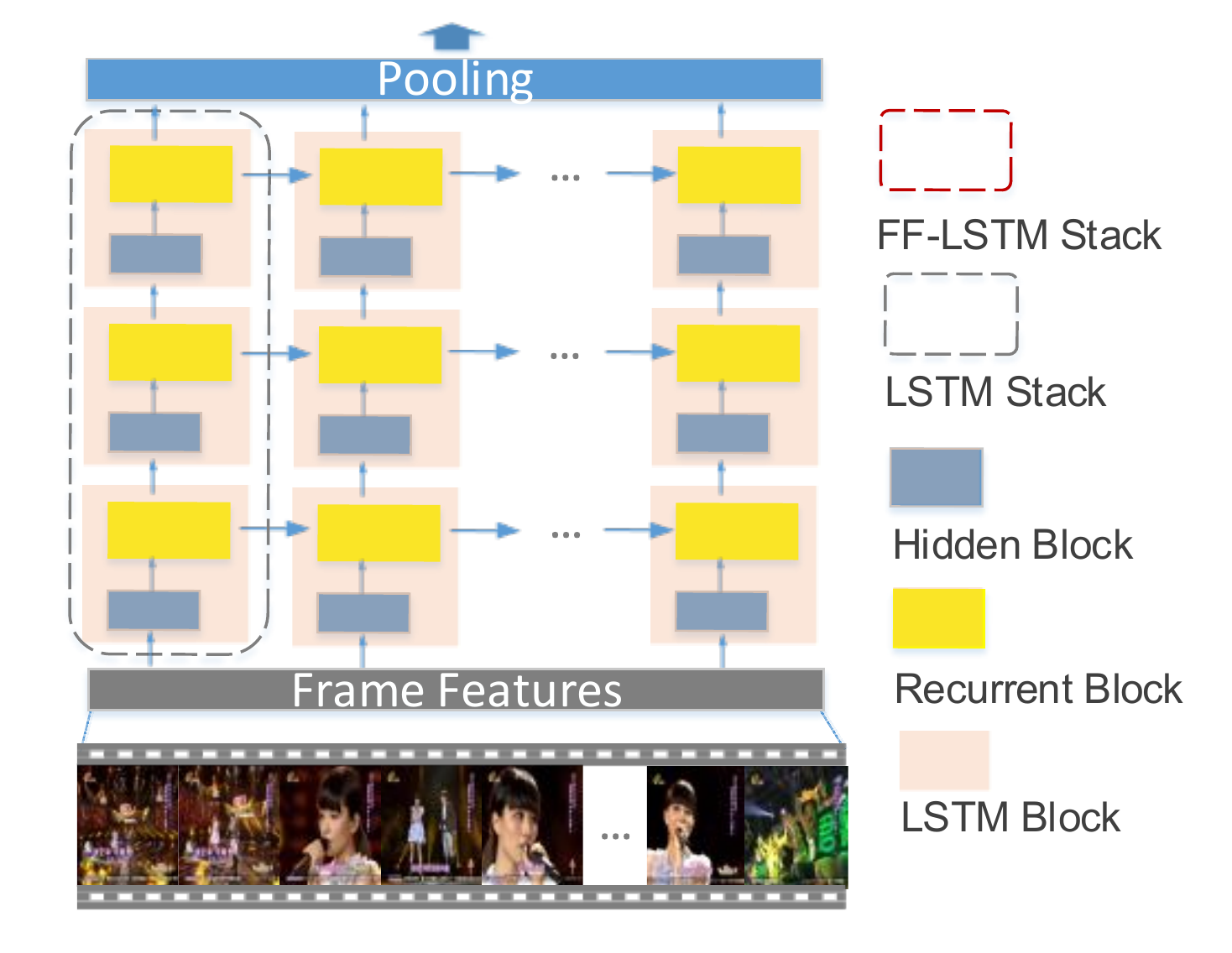}}
\end{center}
\caption{A comparison between FF-LSTM (a) and traditional LSTM (b).}
\label{fig:fig2}
\end{figure*}

Figure 6(b) illustrates a deep LSTM with three stacked layers.
As can be seen, the input of the hidden block is the output of the recurrent block at its previous layer.
In FF-LSTM, a fast-forward connection is added to connect two hidden blocks of adjacent layers.
The added connections build a fast path that contains neither non-linear activations nor recurrent computations such that the information or gradients can be propagated easily.
Figure 6(a) illustrates a FF-LSTM with three layers, and the computation of deep FF-LSTM can be expressed as:
\begin{eqnarray}
\hspace{0.4in}(h^t_k, c^t_k) = {\rm LSTM}_k(f^t_k, h^{t-1}_k, c^{t-1}_k)   \nonumber \\
\hspace{0.4in}f^t_k = W_f^kx^t, \hspace{0.2in}   k = 1 \hspace{0.2in}  \nonumber \\
f^t_k = W_f^k[h^t_{k-1},f^t_{k-1}], \hspace{0.2in} k > 1. \hspace{0.16in}
\end{eqnarray}

Supposing the multi-layer FF-LSTMs receive a proposal range $\{p_{i,s}, p_{i,e}\}$ from video $V_i$, the hidden block of the first FF-LSTM can be calculated as $\{f_1^t = W_f^1u_i^{t+p_{i,s}}\}_{t=0}^{p_{i,e}-p_{i,s}}$.
On the topmost FF-LSTM, a max-pooling layer is used to obtain a global representation of the proposal.
A binary classifier is calculated based on the global representation for story/background classification.
We calculate the intersection-over-union (IoU) between each proposal and ground-truth story, and if the max IoU is larger than 0.7 the proposal is regarded as a positive sample, and a negative sample with IoU less than 0.3.
A boundary regressor is also computed based on the max-pooled global representation.
The multi-task loss over an training proposal $p_{i,k} = \{p_{i,k,s}, p_{i,k,e}\}$ can be written as:
\begin{eqnarray}
\begin{aligned}
L(p_{i,k}) =  L_{cla}(p_{i,k}, c_{i,k}) + \\ \lambda\cdot 1_{c_{i,k}=1} L_{reg}(p_{i,k}, \{s_{i,k},e_{i,k}\} ).
\end{aligned}
\end{eqnarray}

The first term is a cross-entropy loss
\begin{eqnarray}
L_{cla}(p_{i,k}, c_{i,k}) = -\log P(c_{i,k}|p_{i,k}),
\end{eqnarray}
where $c_{i,k}$ is the label of the proposal and $P(c_{i,k}|p_{i,k})$ is the classification score defined by multi-layer FF-LSTMs, max-pooling and the binary classifier.
$\lambda$ is a balanced weight parameter, and $1_{c_{i,k}=1}$ means that the second term works only when the label of the proposal is 1.

The second term accumulates two smooth L1 losses
\begin{eqnarray}
	\begin{aligned}
		L_{reg}(p_{i,k}, \{s_{i,k},e_{i,k}\} ) = {\rm smooth}_{L1}(p_{i,k,s} - s_{i,k}) +\\ {\rm smooth}_{L1}(p_{i,k,e} - e_{i,k}),
	\end{aligned}
\end{eqnarray}
where the definition of ${\rm smooth}_{L1}$ is:
\begin{equation}
{\rm smooth}_{L1}(x)= \left \{\begin{array}{lr}
0.5x^2      & \hspace{0.3in}{\rm if} |x| < 1 \\
|x|-0.5     & \hspace{0.3in}{\rm otherwise}.
\end{array}\right.
\end{equation}
$s_{i,k}$ and $e_{i,k}$ are the beginning and ending indexes of the chosen (with max IoU) ground-truth story of $p_{i,k}$.

\section{Experiments}

\begin{figure}[t]
	\begin{center}
		\includegraphics[scale = 0.53]{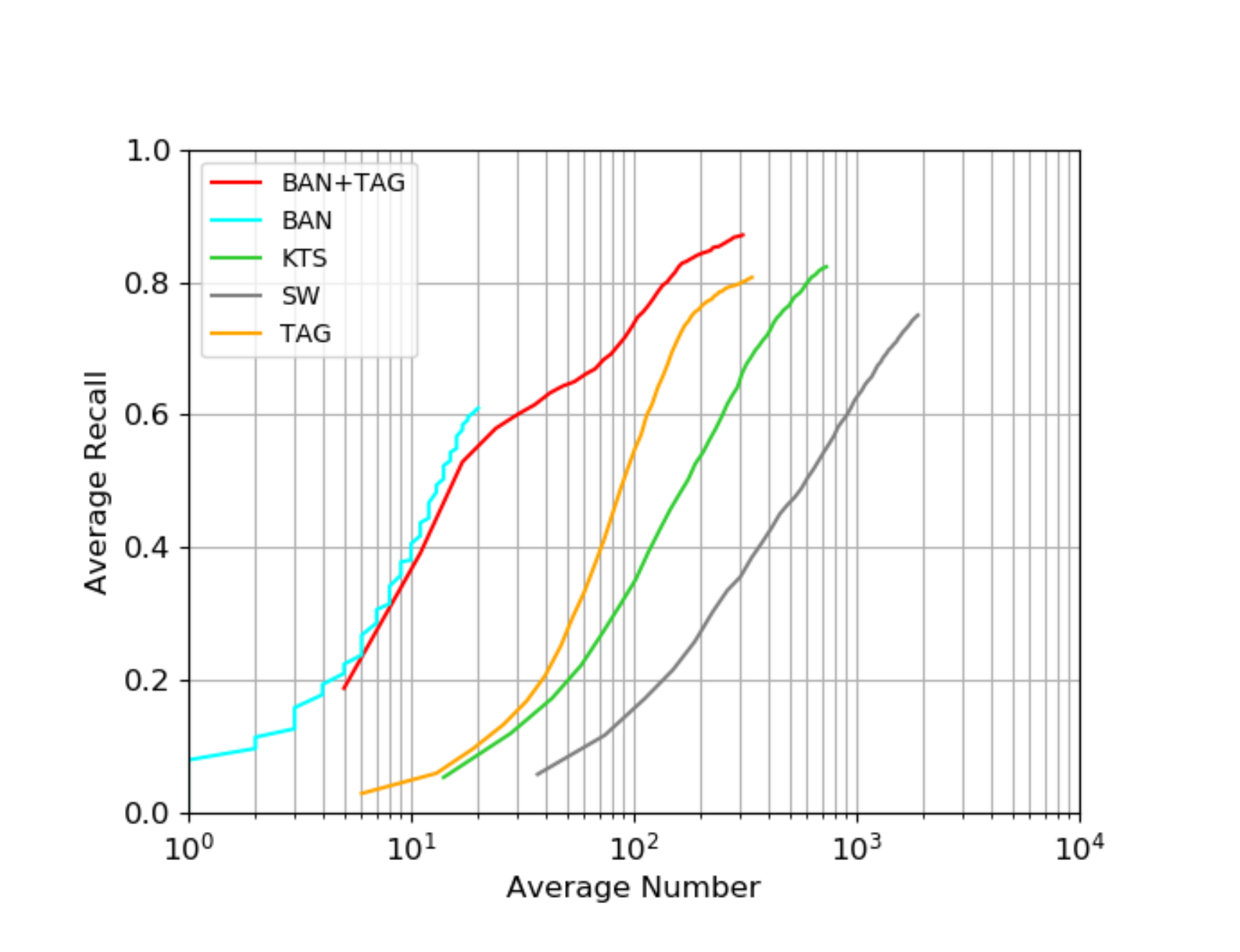}
	\end{center}
	\vspace{-20pt}
	\caption{The AR-AN curves of different temporal proposal generation methods on the TruNet dataset. The average recall of proposed BAN is much higher than other methods when the proposal number is small.
	}\label{fig:fig4}
\end{figure}

\subsection{Implementation Details}
Because the size of the dataset is too large to process, we pre-process the videos and extract frame-level features.
We decode the videos in 1 frame-per-second (FPS) and extract two kinds of features:
``pool5'' of ResNet-50 \cite{he2016resnet} trained on ImageNet \cite{russakovsky2015imagenet} and convolutional audio feature \cite{hershey2017audio}.

We train the BAN using Stochastic Gradient Decent (SGD) with momentum of 0.9, epoch number of 70, weight decay of 0.0005 and a mini-batch size of 256 on four K40 GPUs.
One epoch means all training samples are passed through once.
All parameters are randomly initialized.
The learning rate is set at 0.001.
We tried reducing the learning rate during training but found no benefit.
The sampling ratio of the four categories (within story, background, beginning boundary and ending boundary) is $6:6:1:1$.
To enable boundary regression in the temporal structure modeling stage, we augment the BAN-generated proposals to extend the beginning and ending boundaries similar with \cite{zhao2017temporal}.

We train a 5 layer FF-LSTM using SGD with momentum of 0.9, epoch number of 40, weight decay of 0.0008 and a mini-batch size of 256. The learning rate is kept 0.001 throughout the training.
Positive and negative proposals are sampled with the ratio of $1:3$.
The balanced parameter $\lambda$ is set at 5.

\subsection{Evaluation Metrics}
For story localization, the mean average precision (mAP) of different methods at three different IoU thresholds $\{0.5, 0.7, 0.9\}$ are reported.
We also report the average of mAP with thresholds $[0.5:0.05:0.95]$.
For evaluating the quality of generated temporal proposals, we report the average recall vs. average number of retrieved proposals (AR-AN) curve defined in \cite{gao2017turn}.

\subsection{Ablation Studies}
To study the effectiveness of the proposed BAN, we compare with sliding window search (SW), KTS \cite{potapov2014skim} and TAG \cite{xiong2017tag}.
The comparison results are summarized in Figure 6.
We can see that the average recall of the proposed BAN is significantly higher than other methods when the proposal number is small,
but it cannot generate as much proposals as others because of its selection standard.
Noticing that TAG and BAN are highly complementary, merging their proposals obtains an substantial better curve.

Table 2 summarizes the ablation study results of temporal proposal generation and sequential structure modeling, emerging that both components are crucial for the final performance.
FF-LSTM keeps beating LSTM with different proposal generation methods.
We observe that methods with a training step (TAG and BAN) generate much better proposals than heuristic ones (sliding window and KTS).
Using individual BAN proposals are slight better than using individual TAG proposals, but considering TAG can generate larger number of proposals, BAN and TAG are highly complementary.
As shown in Table 2, merging them brings an obvious improvement over each individual one.

\begin{table}[htb]
	\begin{center}
		\begin{tabular}
			{c||c|c}\hline
			&  LSTM &  FF-LSTM\\\hline\hline
			SW & 40.63 & 44.34 \\\hline
			KTS &  41.08 & 48.03   \\\hline
			BAN & 51.92 & 55.80  \\\hline
			TAG &   51.15 &54.03  \\\hline
			TAG + BAN &  52.55 & \textbf{57.34} \\\hline
		\end{tabular}
		
	\end{center}
	\caption{Ablation study of temporal proposal generation and sequential structure modeling. SW refers to sliding window, KTS refers to kernel temporal segmentation \cite{potapov2014skim}, TAG refers to temporal actionness grouping \cite{xiong2017tag}. The reported number is the average mAP with different thresholds. }
\end{table}

Table 3 summarizes the ablation study results of using different features.
As can be seen, the RGB and audio features are highly complementary and dropping either feature decreases the performance obviously.
Another interesting phenomenon is using single audio feature achieves better performance than using single RGB feature.
We guess this is because audio patterns change rapidly in video story boundaries, especially for variety programs.

\begin{table}[htb]
	\begin{center}
		\begin{tabular}
			{c||c}\hline
			& mAP \\\hline
			RGB & 41.33 \\\hline
			Audio & 51.0 \\\hline
			RGB + Audio & \textbf{57.34} \\\hline
		\end{tabular}
		
	\end{center}
	\caption{Ablation study of feature combinations. }
\end{table}

Table 4 summarizes the ablation study results of using different deep recurrent models for sequential structure modeling.
Directly stacking multiple LSTM layers lead to performance drop because of convergence difficulty.
Instead, using deeper FF-LSTM obtains noteworthy performance gains.
A 3-layer FF-LSTM increases the mAP by 3.9 over 1-layer LSTM, and a 5-layer FF-LSTM furter increases the mAP by 0.9.
We also tried more than 5-layer FF-LSTM, but found very marginal improvements.

\begin{table}[htb]
	\begin{center}
		\begin{tabular}
			{c||c}\hline
			& mAP \\\hline
			LSTM &   52.55 \\\hline
			3-layer LSTM & 46.42  \\\hline
			5-layer LSTM & 50.57 \\\hline
			3-layer FF-LSTM  &  56.41  \\\hline
			5-layer FF-LSTM & \textbf{57.34}    \\\hline
		\end{tabular}
		
	\end{center}
	\caption{Ablation study of deep sequence models. }
\end{table}

Table 5 summarizes the ablation study results of the regression loss.
From the results, we can see that adding the regression loss improves the mAP by 3.0.
Considering that BAN has addressed the story boundaries when generating proposals,
our FF-LSTM component can capture high-order dependencies among frames for further boundary refinement.

\begin{table}[htb]
	\begin{center}
		\begin{tabular}
			{c||c}\hline
			& mAP \\\hline
			Cla. Loss &  53.37 \\\hline
			Cla. Loss + Reg. Loss & \textbf{57.34} \\\hline
		\end{tabular}
	\end{center}
	\caption{Ablation study of regression loss. }
	
\end{table}

\subsection{Quantitative Comparison Results}
We compare our proposed framework with state-of-the-art video summary methods, including vsLSTM \cite{zhang2016summarization} and HD-VS \cite{yao2016highlight}.
We have to re-implement vsLSTM and HD-VS to make them suitable for our formulation,
because the completeness of the storytelling is essential in story-preserving long video truncation, but is not considered in previous video summary papers.

For vsLSTM, we use the merged TAG and BAN proposals as the input of vsLSTM, which replaces the FF-LSTM and servers as the basis of temporal modeling for regression and classification.
For HD-VS, we use the merged TAG and BAN proposals as the input of a 5-layer FF-LSTM, and the cross-entropy loss is replaced with a deep ranking loss proposed in HD-VS.
The results are summarized in Table 6.
As can be seen, the proposed framework outperforms all previous methods in all cases in a large margin.

\begin{table}[htb]
	\begin{center}
		\begin{tabular}
			{c|c|c|c||c}\hline
			\multicolumn{5}{c} {\textbf{mAP@$\alpha$}} \\\hline
			& 0.5 & 0.7 & 0.9 & Average\\\hline
			vsLSTM & 69.31 & 61.01 & 34.03 & 53.0 \\\hline
			HD-VS& 63.89 & 49.52 & 15.70 & 41.22 \\\hline
			Ours & \textbf{71.80} & \textbf{65.45} & \textbf{39.24} & \textbf{57.34} \\\hline
		\end{tabular}
	\end{center}
	
	\caption{Comparison results measured by mAP at different IoU thresholds $\alpha$. The average mAPs in different thresholds are also reported.}
\end{table}

\subsection{User-Study Results}
We finally conduct subjective evaluation to compare the quality of generated short video story summary.
100 volunteers with different genders, education backgrounds and ages are required to process the following steps independently:
\begin{itemize}
\item Watch 18 randomly selected long videos.
\item For each long video, watch the short video story summaries generated by the proposed framework, vsLSTM \cite{zhang2016summarization} and HD-VS \cite{yao2016highlight}.
\item For each long video, choose one of the video story summaries as the best one.
\end{itemize}

The answers of the 100 volunteers are accumulated and the chosen ratio of the three methods are summarized in Table 7.
As can be seen, the proposed framework generated summaries receive 46.4\% of the votes, which is higher than vsLSTM (40.8\%) and HD-VS(12.8\%).

\begin{table}[htb]
	\begin{center}
		\begin{tabular}
			{c||c}\hline
			& Chosen Ratio \\\hline
			Ours & 46.4\% \\\hline
			vsLSTM \cite{zhang2016summarization} & 40.8\% \\\hline
			HD-VS \cite{yao2016highlight} & 12.8\% \\\hline
		\end{tabular}
		
	\end{center}
	\caption{Ablation study of feature combinations. }
\end{table}

\section{Conclusions}
In this paper, we propose a new story-preserving video truncation problem that requires algorithms to truncate long videos into short, attractive, and unbroken stories. This problem is particularly important for resource production in video sharing platforms.
We collect and annotate a new large TruNet dataset and propose a novel framework that combines BAN and FF-LSTM is proposed to address this problem.

\bibliographystyle{ieee_fullname}
\bibliography{highlight}
\end{document}